\documentclass[conference]{IEEEtran}

\IEEEoverridecommandlockouts
\usepackage[numbers,sort&compress]{natbib}
\usepackage{amsmath,amssymb,amsfonts}
\usepackage{algorithmic}
\usepackage{graphicx}
\usepackage{textcomp}
\usepackage{booktabs} 
\usepackage{array}  
\usepackage{xcolor}
\usepackage{tabularx} 
\usepackage{multicol} 
\usepackage{fontawesome}
\usepackage[utf8]{inputenc}
\usepackage{amsmath}
\usepackage[colorlinks=true, linkcolor=blue, citecolor=blue]{hyperref}
\usepackage{placeins} 
\def\BibTeX{{\rm B\kern-.05em{\sc i\kern-.025em b}\kern-.08em
  T\kern-.1667em\lower.7ex\hbox{E}\kern-.125emX}}
\usepackage{orcidlink}
\DeclareRobustCommand*{\IEEEauthorrefmark}[1]{%
	\raisebox{0pt}[0pt][0pt]{\textsuperscript{\footnotesize\ensuremath{#1}}}}
	
\begin{document}
\title{DAGNet: A Dual-View Attention-Guided Network for Efficient X-ray Security Inspection \\
\thanks{\textsuperscript{\faEnvelopeO} Corresponding author. }
\thanks{This work was supported in part by the National Natural Science Foundation of China under Grant 62171141, and in part by Guangdong Science and Technology Department under Grant 2024A1515011803.}
}
\author{
	\IEEEauthorblockN{
		Shilong Hong\IEEEauthorrefmark{1},
		Yanzhou Zhou\IEEEauthorrefmark{1, 2},
		Weichao Xu\IEEEauthorrefmark{1}\textsuperscript{\faEnvelopeO}
	}
	\IEEEauthorblockA{\IEEEauthorrefmark{1}School of Automation, Guangdong University of Technology, Guangzhou 510006, China}
	\IEEEauthorblockA{\IEEEauthorrefmark{2}Guangzhou Institute of Science and Technology, Guangzhou 510006, China}
	
	\IEEEauthorblockA{panda@pandacow.cn, \{zhouyanzhou, wcxu\}@gdut.edu.cn}
}

\maketitle
\begin{abstract}
With the rapid development of modern transportation systems and the exponential growth of logistics volumes, intelligent X-ray-based security inspection systems play a crucial role in public safety. Although single-view  X-ray baggage scanner is widely deployed, they struggles to accurately identify contraband in complex stacking scenarios due to strong viewpoint dependency and inadequate feature representation. To address this, we propose a Dual-View Attention-Guided Network for Efficient X-ray Security Inspection (DAGNet).
This study builds on a shared-weight backbone network as the foundation and constructs three key modules that work together: 
(1) Frequency Domain Interaction Module (FDIM) dynamically enhances features by adjusting frequency components based on inter-view relationships;
(2) Dual-View Hierarchical Enhancement Module (DVHEM) employs cross-attention to align features between views and capture hierarchical associations;
(3) Convolutional Guided Fusion Module (CGFM) fuses features to suppress redundancy while retaining critical discriminative information.
Collectively, these modules substantially improve the performance of dual-view X-ray security inspection.
Experimental results demonstrate that DAGNet outperforms existing state-of-the-art approaches across multiple backbone architectures.
The code is available at: \href{https://github.com/ShilongHong/DAGNet.}{https://github.com/ShilongHong/DAGNet.}
\end{abstract}

\begin{IEEEkeywords}
	Dual-view X-ray inspection, feature fusion, frequency domain, multi-scale, cross-view attention, convolutional attention, X-ray security inspection.
\end{IEEEkeywords}

\section{Introduction}
As modern transportation systems develop rapidly and logistics volumes grow exponentially, public security faces increasingly severe challenges\cite{akcay2017evaluation}. 
In this context, X-ray security inspection systems play a crucial role in detecting prohibited items and ensuring public safety. 
X-ray inspection equipment is widely used in airports, train stations, customs, post offices, and other public places to examine luggage, parcels, and other items, ensuring they do not contain dangerous goods, contraband, or explosives. However, due to the growing complexity of transportation systems, traditional single-view X-ray baggage scanner has become inadequate for meeting rising security demands.

The widespread use of single-view X-ray baggage scanner has provided basic security guarantees but still faces inherent problems. Due to the limitations of single-view imaging principles, X-rays pass through packages, and materials of varying compositions, densities, and thicknesses absorb X-rays differently, affecting the display of images. In complex stacking scenarios, single-view systems often fail to capture complete information about overlapping items, significantly compromising the accuracy of detecting prohibited objects.\cite{9710407, 9722843, akcay2018using, miao2019sixray, tao2021towards}. 

\begin{figure}[t]
	\includegraphics[width=0.48\textwidth]{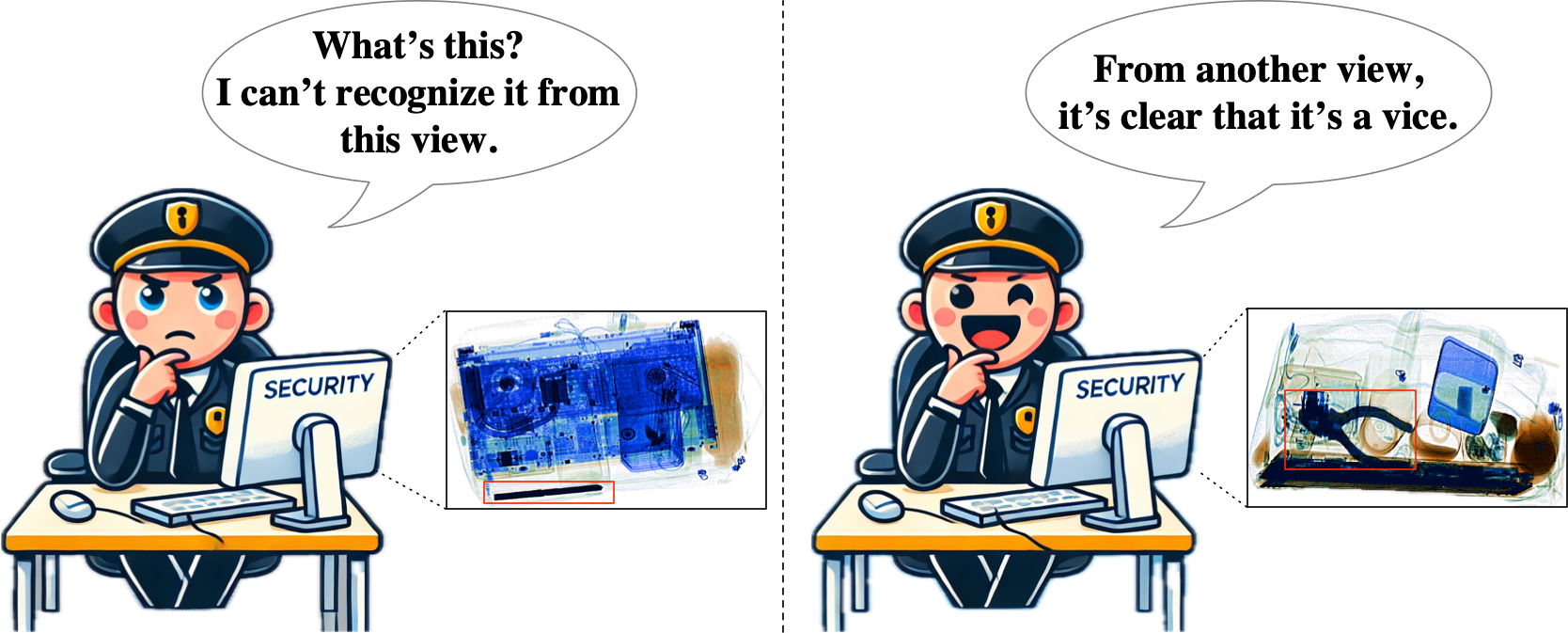}
	\caption{Overall architecture of the proposed framework. }
	\label{fig:dual}
\end{figure}

With technological advancements, dual-view X-ray inspection technology has gradually become an important means to address these problems. 
By imaging objects simultaneously from two orthogonal directions, dual-view X-ray technology greatly improves spatial representation. 
This is particularly beneficial in complex stacking, angular deviations, and object occlusions, as it can provide more complete contours and morphological information of prohibited items, thereby improving detection accuracy\cite{ma2024towards}. 
Although these methods can alleviate the viewpoint difference problem and supplement the information to a certain extent, it is difficult to fully exploit the detail complementarity between viewpoints, which makes the information redundancy and lack of details due to the viewpoint difference, and ultimately leads to the detection effect being adversely affected.

\begin{figure*}[t]
	\includegraphics[width=\textwidth]{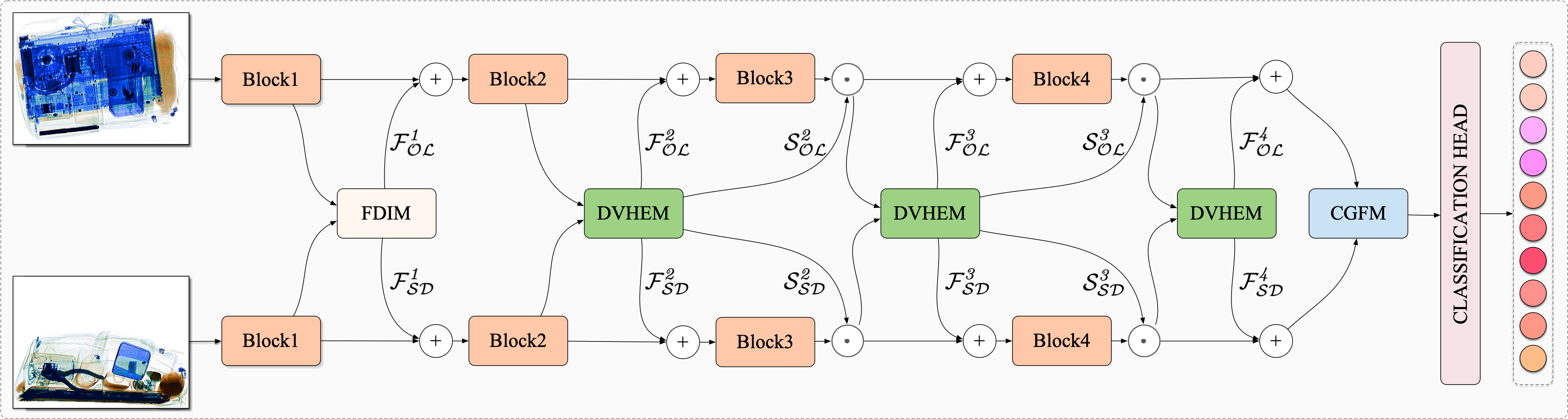}
	\caption{Overall architecture of the proposed framework. }
	\label{fig:all}
\end{figure*}

To address these challenges, we propose DAGNet, an attention-guided network designed specifically for dual-view X-ray security screening. We formulate the task as a multi-label, multi-category classification problem, effectively overcoming geometric variations and modality-specific differences inherent in dual-view imaging.
This task definition can effectively avoid the limitations of the detection paradigm due to cross-view geometric variations and modality-specific differences, making the model more focused on attribute-level feature associations between views, and has become one of the mainstream solutions in current related research. Different from the existing approaches, the main contribution of this study lies in the design of a systematic multi-level feature interaction architecture, which effectively solves the problem of information loss in dual-view fusion.

In the design of the model, this study takes the backbone network with shared weights as the basis and constructs three mutually synergistic key modules as the main innovations:
\begin{itemize}
	\item The Frequency Domain Interaction Module (FDIM), dynamically enhancing features by adjusting frequency components based on inter-view relationships
	\item The Dual-View Hierarchical Enhancement Module (DVHEM), employing cross-attention mechanisms to adaptively align and integrate features across views at multiple scales
	\item The Convolutional Guided Fusion Module (CGFM), efficiently fusing features to suppress redundancy while preserving discriminative information. 
\end{itemize}



\section{Related Research}
In the field of X-ray security image processing, most existing approaches are designed from a single-view perspective, predominantly enhancing feature representations to tackle challenges such as occlusion and complex backgrounds. For instance, the De-Occlusion Aware Attention Module (DOAM)\cite{wei2020occluded} focuses on extracting edge information from images, reducing the interference caused by occluded objects by concentrating on their edge features. CFPA-Net\cite{9754631} employs cross-layer feature fusion to reinforce the interaction between high-level semantic features and low-level location cues, thereby improving detection accuracy in complex backgrounds. IEFPN\cite{9766055} introduces a Layer-based Recalibration Module (LRM) to optimize multi-scale feature fusion on top of a traditional Feature Pyramid Network (FPN), resulting in better detection of targets with varying scales. Methods such as Re-BiFPN\cite{tfa2022rebifpn}, and FDTNet\cite{ZHU2024108076} have also proposed innovations in feature pyramids, attention mechanisms, and dual-stream networks (combining frequency and spatial domains), respectively, and have seen notable performance gains. 
However, these single-view methods are constrained by the lack of complementary viewpoint information, leading to difficulties in addressing severe occlusions and overlapping objects from extreme viewing angles.

These methods lay a crucial foundation for subsequent research and collectively improve the performance of prohibited object detection. However, they still face challenges, particularly in effectively using frequency domain and edge information to address occlusion issues.

To further overcome the limitations of single-view methods, researchers have begun exploring dual-view data and networks\cite{tao2024dual}. With the introduction of dual-view datasets, the Attentive Hierarchical Cross-View Representation (AHCR)\cite{ma2024towards} approach leverages information exchange between two viewpoints, achieving better detection accuracy in occluded and complex environments. 
Nevertheless, current dual-view methodologies do not adequately leverage the inherent structural characteristics and complementary details of X-ray security images, especially in complex scenarios with extreme occlusions or multiple overlapping items.
Some researchers have proposed Transformer-based dual-view methods\cite{meng2024transformer} by refining the attention mechanism for inter-view information exchange. However, these methods exhibit weak generalizability, making them difficult to adapt across various model architectures.

Based on the above analysis, dual-view methods inherently excel at addressing occlusions and overlapping targets, yet the efficient extraction of complementary information between the two viewpoints—particularly leveraging the structural features in X-ray imaging—remains a pressing challenge. In response, we propose a multi-scale interactive feature fusion framework that fully integrates frequency-domain features and attention mechanisms, thereby enhancing feature representation and target recognition accuracy in dual-view X-ray images.

\section{Methodology}
\subsection{Overall Architecture}

As illustrated in \autoref{fig:all}, the DAGNet proposed in this work aims to maximize feature extraction efficiency and achieve efficient and accurate security inspection tasks. The entire framework is divided into four main blocks based on changes in the size of the feature maps output by the backbone network. Considering that the feature information of the two perspectives partially overlaps and complements each other, this study adopts a backbone network structure with shared weights to ensure that dual-perspective feature extraction remains efficient while allowing the backbone network to use as much data as possible, making the model more robust and consistent. The specific process is as follows.

First, given a pair of dual-view images, two shared backbone capture features from different viewpoints. After the first backbone network module, the feature maps are input into {FDIM}, where Fourier transform is used to extract frequency domain features and establish inter-viewpoint frequency domain information associations in the frequency domain.

Next, the feature maps output from the backbone of different perspectives are sent to {DVHEM} for feature interaction between perspectives. Adaptive multi-scale feature enhancement weights are innovatively introduced to further improve the model's ability to distinguish targets in complex scenes.

Finally, the high-level features obtained from the two perspectives are sent to {CGFM} for perspective feature fusion, and the fused features are sent to the classification head to complete the final output.

In the subsequent sections, we will elaborate on the detailed structure and operational principles of each proposed module.
\subsection{Frequency Domain Interaction Module (FDIM)}
\begin{figure}[t]
	\centering
	\includegraphics[width=0.48\textwidth]{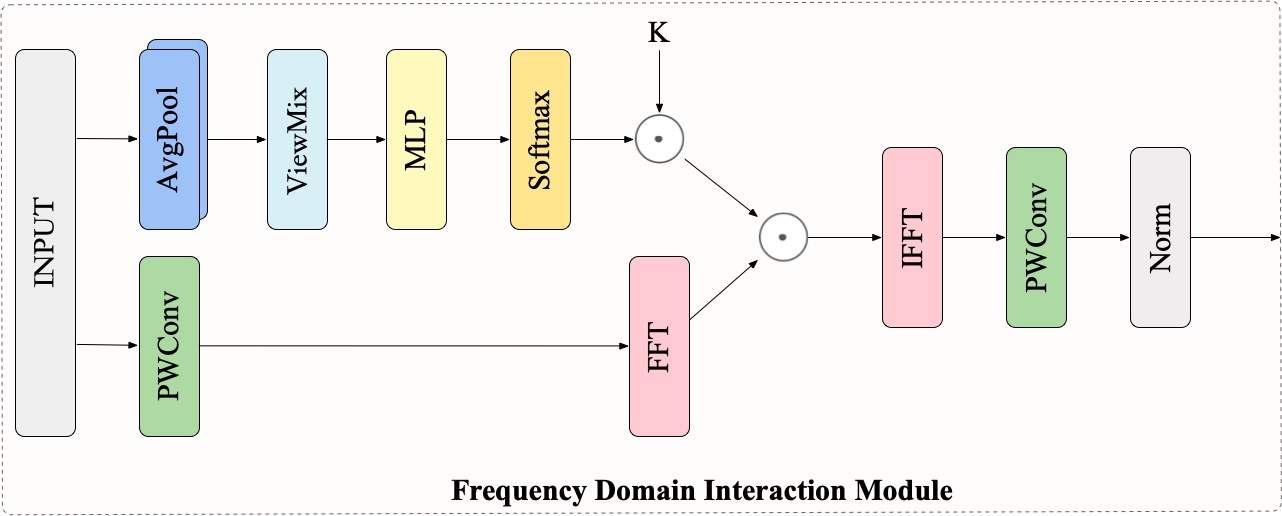}
	\caption{The overall architecture of the proposed Frequency Domain Interaction Module (FDIM)}
	\label{fig:fdim}
\end{figure}

Capturing features of overlapping items is challenging in dual-view X-ray imaging due to its transmissive imaging nature. To effectively address this issue, we propose the Frequency Domain Interaction Module (FDIM), illustrated in \autoref{fig:fdim}, leveraging frequency-domain analysis to enhance feature extraction capabilities.

Unlike previous frequency domain methods, FDIM dynamically adjusts and enhances frequency components by collaboratively constructing adaptive filters from dual-view features. Specifically, important frequency bands are amplified while noisy bands are suppressed, effectively leveraging global and frequency-selective properties to yield more discriminative features compared to spatial-domain approaches.

FDIM first performs Pointwise Convolution (PWConv) \cite{zhang2020high} on the input feature map \(x\) to integrate channel information, and then applies a two-dimensional fast Fourier transform\cite{gonzalez2009digital} to map it from the spatial domain to the frequency domain:
\begin{equation}
	{F_X}' = \mathit{FFT}({PWConv(x)})
	\label{eq:fdim_fft}
\end{equation}
where $\mathit{x}$ is the input feature map, $\mathit{PWConv}(\cdot)$ represents  the pointwise convolution operation, $\mathit{FFT}(\cdot)$ denotes the 2D Fast Fourier Transform (FFT) , and the output $\mathit{F_X}'$ is the feature map in the frequency domain.

Subsequently, to determine the appropriate filter parameters, inspired by Tatsunami et al.\cite{tatsunami2024fft} in their work on frequency domain, this design divides the filter D into two parts: a learnable base filter and a channel-wise weighted weight matrix to better enhance feature expression.

First, the filter weight matrix $W_x$  is generated through interactions between dual-view input features, constructing a learnable channel weighting mechanism that emphasizes key channel responses. This is formally defined as:
\begin{equation}
	x' = P_m(x) \cdot ViewMix(x,y) + P_m(x) 
\end{equation}
\begin{equation}
	ViewMix(x,y) = Sigmoid(PWConv(P_m(y)))
\end{equation}
\begin{equation}
	W_x = \mathit{SoftMax(MLP(x'))}
	\label{eq:fdim_global_mean}
\end{equation}
where $P_m(x)$ denotes global average pooling on the input feature map, $x$, $y$ represents the input feature map from another view, $ViewMix(x,y)$ denotes the interaction between features from two views, $\mathit{PWConv}(\cdot)$ denotes channel-wise convolution, $\mathit{MLP}(\cdot)$ is a multi-layer perceptron, and $\mathit{Sigmoid}(\cdot)$ denotes the sigmoid function. $\mathit{SoftMax}(\cdot)$ is the normalized exponential function, and $W_x$ denotes the final weight matrix.

Subsequently, the learned base filter $K$ is modulated at the channel level to produce the adaptive frequency-domain filter $\mathit{D_X}$:

\begin{equation}
	\mathit{D_X} = \mathit{{K}} \cdot W_x
	\label{eq:fdim_dynamic_filtering}
\end{equation}
where $\mathit{D_X}$ is the generated frequency domain filter, $\mathit{K}$ is the learnable basic filter, a complex matrix with dimensions $H \times \lfloor W/2 \rfloor$, and $W_x$ is the channel-level weight matrix used to adjust the response intensity of different channels in the frequency domain.

The enhanced frequency-domain feature map is then transformed back to the spatial domain using inverse FFT and integrated through channel-wise convolution. Finally, residual connections stabilize feature learning:
\begin{equation}
	F_X = PWConv(\mathit{F}^{-1}(\mathit{{F_X}}' \cdot \mathit{D_X})) + x
	\label{eq:fdim_merge_x}
\end{equation}
where, $\mathit{F}^{-1}(\cdot)$ denotes the two-dimensional inverse fast Fourier transform, $\mathit{PWConv}(\cdot)$ denotes the channel-wise convolution operation, $F_X$ is the original input feature map, $\mathit{D_X}$ is the generated frequency domain filter, and the final output $F_X$ is the feature map after fusion of the frequency domain enhancement information.

\subsection{Dual-View Hierarchical Enhancement Module (DVHEM)}
\begin{figure}[t]
	\centering
	\includegraphics[width=0.48\textwidth]{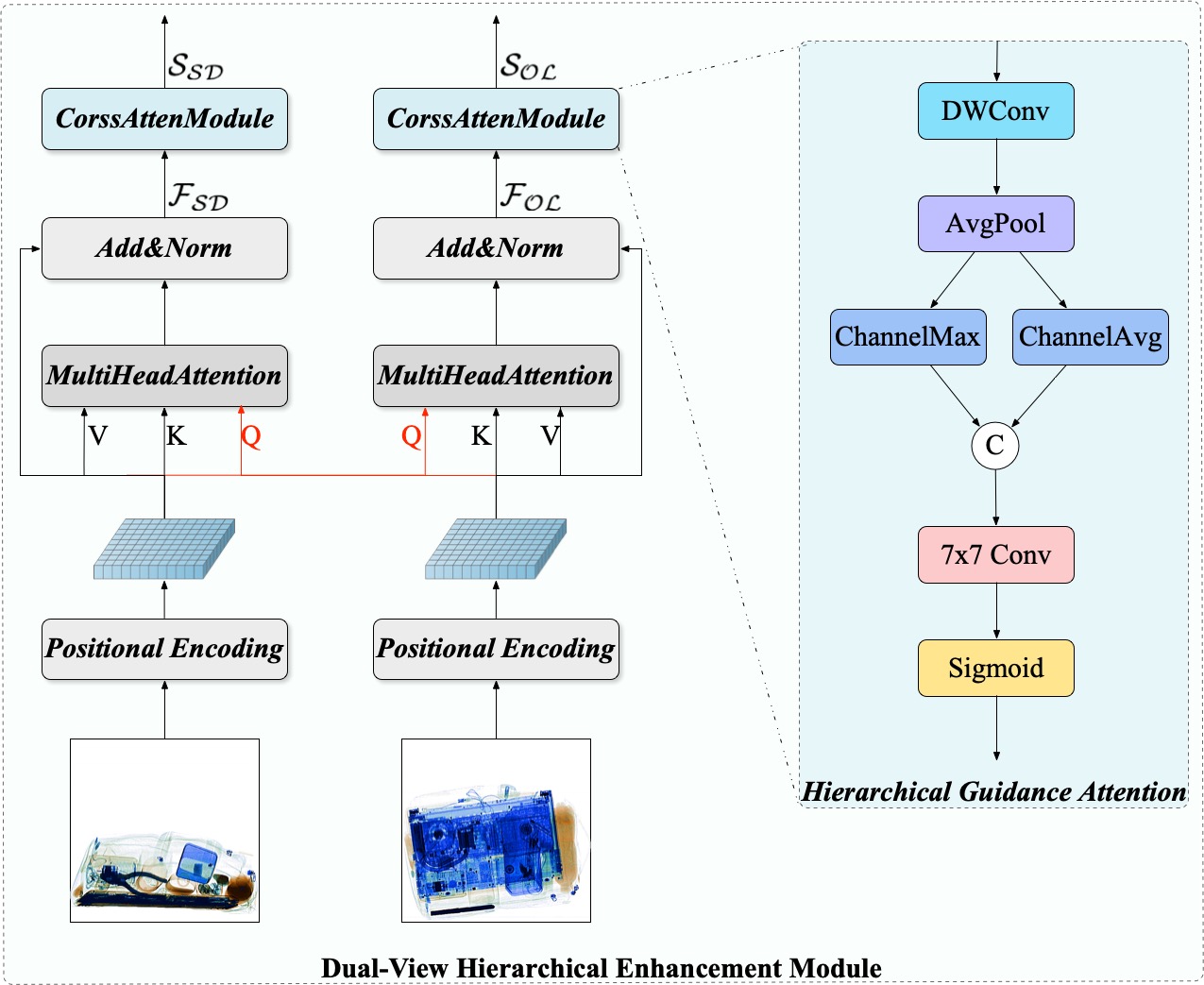}
    \caption{Overall architecture of the Dual-View Hierarchical Enhancement Module (DVHEM).}  	
	\label{fig:DVHEM}
\end{figure}

In security inspection scenarios, dual-view X-ray images exhibit distinct feature distributions. Although the same object exhibits feature differences under different views, it still contains significant associated information. Additionally, due to the differing shapes and sizes of features presented in different views, there exists potential associativity between low-level and high-level features. To fully leverage these characteristics, this work proposes the DVHEM, as shown in \autoref{fig:DVHEM}. The module consists of two key components: first, it enhances the feature representations of each view through interactions based on the relationships between views; second, it constructs a Hierarchical Guidance Attention module (HGA) that guides attention across levels.
This design fully integrates and utilizes the correlations between features from different viewpoints and scales, balancing global information and local details, thereby significantly improving performance in dual-view security inspection scenarios.

First, DVHEM adds position encoding to the input features of the vertical viewpoint branch \(F_{OL}\) and the horizontal viewpoint branch \(F_{SD}\) to introduce spatial position information, as defined below:
\begin{equation}
	F_{OL}^{P} = F_{OL} + {PosEnc}(F_{OL})
	\label{eq:DVHEM_pos_OL}
\end{equation}
\begin{equation}
	F_{SD}^{P} = F_{SD} + {PosEnc}(F_{SD})
	\label{eq:DVHEM_pos_SD}
\end{equation}
Here, $F_{OL}$ and $F_{SD}$ denote the input feature maps of the vertical and horizontal view angle branches, respectively, ${PosEnc}(\cdot)$ denotes the cosine position encoding function, and $F_{OL}^{P}$ and $F_{SD}^{P}$ denote the feature maps after adding position encoding.

Next, DVHEM adopts a multi-head cross-attention mechanism\cite{fang2025guided,dosovitskiy2020image, liu2021swin} to promote effective feature interaction between vertical and horizontal viewpoints. In this mechanism, the features of one branch are used as Q (Query), and the features of the other branch are used as K (Key) and V (Value), which can be defined as:
\begin{equation}
	F_{SD} = MHA(Q=F_{OL}^{P}, K=F_{SD}^{P}, V=F_{SD}^{P})
\end{equation}
\begin{equation}
	F_{OL} = MHA(Q=F_{SD}^{P}, K=F_{OL}^{P}, V=F_{OL}^{P})
\end{equation}
where $MHA(\cdot)$ denotes the multi-head attention mechanism, $Q$ , $K$ , and $V$ are the query, key, and value in the attention mechanism, respectively.

To further enhance information transmission between different layers, the feature maps obtained from cross-attention are input into the proposed HGA module. First, this module uses Depthwise Separable Convolution (DWConv) \cite{howard2017mobilenets, chollet2017xception} for preliminary feature extraction and dimension reduction to reduce the computational burden while highlighting feature regions, as expressed below:
\begin{equation}
	F_{OL}^{down} = {AvgPool}({DWConv}(F_{OL}))
\end{equation}
where $F_{OL}$ is the vertical view angle branch feature map, $DWConv(\cdot)$ denotes depth-separable convolution, $AvgPool(\cdot)$ denotes average pooling, which reduces the feature map resolution and enhances regional aggregation capabilities, and $F_{OL}^{down}$ denotes the processed intermediate features.

Next, channel-dimensional information extraction and enhancement are performed on the feature maps to capture a more comprehensive feature representation, as follows:
\begin{equation}
	F_{OL}^{Channel} = Concat(C_M(F_{OL}^{down}), C_A(F_{OL}^{down}))
\end{equation}
where $C_M(\cdot)$ and $C_A(\cdot)$ denote the maximum pooling (ChannelMax) and average pooling (ChannelAvg) operations in the channel dimension, respectively, and $Concat(\cdot)$ denotes the feature concatenation operation. The output $F_{OL}^{Channel}$ is the feature map enhanced at the channel level.

Finally, a 7×7 convolution is used to further enhance the spatial feature representation capability, and the output is mapped to the range (0, 1) through the Sigmoid activation function to generate the guidance attention weight maps \(S_{OL}\) and \(S_{SD}\), which are then passed to the next layer through matrix multiplication. This ensures that the model can obtain more global structural information and image details. Taking the generation of \(S_{OL}\) as an example, this process can be expressed as:
\begin{equation}
	S_{OL} = Sigmoid(Conv_7(F_{OL}^{Channel}))
\end{equation}
where $Conv_7(\cdot)$ denotes a convolution operation with a kernel size of $7\times7$, $Sigmoid(\cdot)$ denotes the Sigmoid activation function, and $S_{OL}$ denotes the generated guidance attention weight map.

\subsection{Convolutional Guided Fusion Module (CGFM)}
Directly concatenating or simply adding together feature maps from different sources often introduces redundant information or noise, thereby reducing the discriminative ability of the features. 
To solve this problem and efficiently fuse heterogeneous features from two perspectives, this study proposes CGFM, which combines spatial–channel attention with a dual-branch bottleneck fusion to produce highly discriminative features.

\begin{figure}[t]
	\centering
	\includegraphics[width=0.48\textwidth]{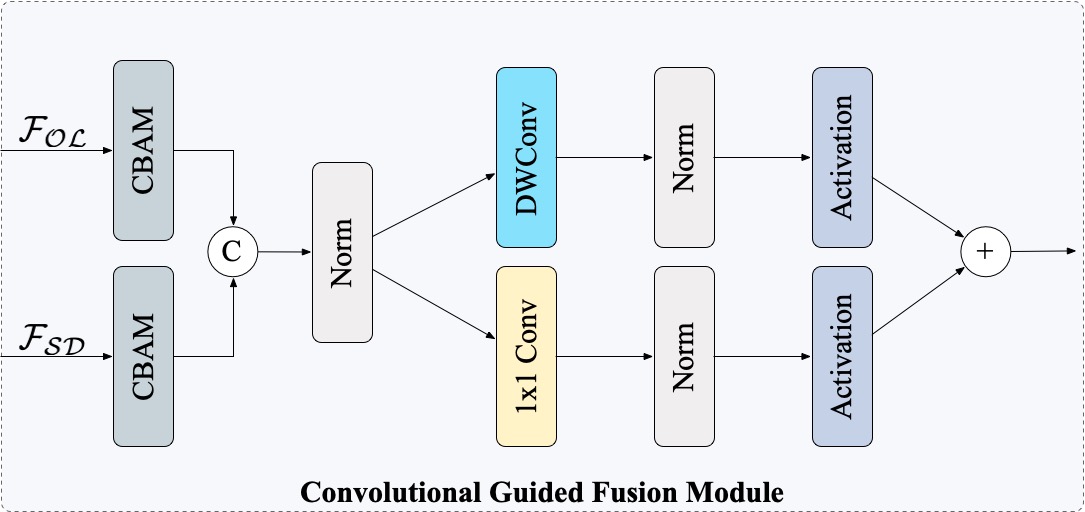}
	\caption{The overall architecture of the proposed Convolutional Attention Fusion Module (CGFM)}
	\label{fig:CGFM}
\end{figure}

As shown in \autoref{fig:CGFM},
this module introduces a Convolutional Block Attention Module (CBAM\cite{woo2018cbam}) and Depthwise Separable Convolution (DWConv) to achieve dual-view feature fusion in an efficient and discriminative manner, thereby enhancing the detection capability of the model.

Formally, let \(F_{OL}\) and \(F_{SD}\) represent the feature maps of the two views. First, the CBAM processes the two feature maps separately, adaptively adjusting the importance of different regions and channels to highlight key features and suppress redundant information:
\begin{equation}
	F_{OL}' = CBAM(F_{OL}), \quad
	F_{SD}' = CBAM(F_{SD})
\end{equation} 
where $F_{OL}$ and $F_{SD}$ represent the original feature maps from vertical and horizontal views, respectively, and $CBAM(\cdot)$ denotes the channel-spatial attention mechanism, which enhances discriminative regional features while suppressing redundant or noisy features, outputting $F_{OL}'$ and $F_{SD}'$ as attention-optimized feature maps.

Subsequently, the optimized feature maps are concatenated along the channel dimension and normalized to generate the preliminary fused feature map $F_{bn}$:
\begin{equation}
	F_{bn} = BatchNorm(Concat(F_{OL}', F_{SD}'))
\end{equation}
where $BatchNorm(\cdot)$ denotes batch normalization, $Concat(\cdot)$ denotes concatenation of attention-enhanced features from two views along the channel dimension, and $F_{bn}$ denotes the normalized fusion feature map.


Next, following the classic bottleneck structure design, the features are sent to two branches for processing: one uses DWConv to retain spatial structural information, and the other uses 1×1 convolution to integrate cross-channel information. Finally, the results of the two branches are summed. This design reduces complexity while balancing the expression of spatial features and channel relationships. The specific process can be represented as follows:
\begin{equation}
	F_{out} = DWConv(F_{bn}) + Conv_1(F_{bn})
	\label{eq:dwcsplayer_sum}
\end{equation}
where $DWConv(\cdot)$ denotes the depth-separable convolution operation, $Conv_1(\cdot)$ denotes the $1 \times 1$ convolution operation, and $F_{out}$ denotes the final output feature map after fusion.

\section{Experiments}
\subsection{Dataset Description}
This study utilizes the DvXray\cite{ma2024towards} dataset provided by Bowen Ma et al., aimed at advancing prohibited item detection in luggage security screening. DvXray is a large-scale dual-view X-ray image dataset comprising 32,000 images, including 5,000 pairs of positive samples and 11,000 pairs of negative samples. The dataset samples are sourced from real subway stations (approximately 66\%) and manually designed luggage scanned using X-ray baggage scanner (approximately 34\%). It includes 15 categories of common prohibited items, such as firearms, knives, and explosives. All images are in PNG format with a resolution of 800$\times$600 pixels and are annotated at the image level.

DvXray is currently the only publicly available dual-view X-ray dataset. Other dual-view datasets are either not publicly released or do not match the scale and category coverage of DvXray. Therefore, DvXray was selected as the primary data source to ensure the reproducibility and reliability of the experiments.

The dual-view design of the DvXray dataset offers a significant advantage in detecting overlapping items, thereby enhancing the accuracy of prohibited item recognition. The annotation process was carried out by an internal team and verified by professional security personnel to ensure accuracy and reliability. While multi-view images might further improve recognition performance, the dual-view approach strikes an effective balance between computational efficiency and detection effectiveness, meeting the requirements of this study.

\begin{table*}[t]
	\centering
	\caption{Comparison of Model Performance Across Different Backbones}
	\label{tab:comparison_experiment}
	\begin{tabularx}{\textwidth}{l *{4}{>{\centering\arraybackslash}X}}
		\toprule
		Model & FLOPs & Params & Val\_mAP & Test\_mAP \\
		\midrule
		ResNet50-Dual\cite{he2016deep} & 10.793G & 23.539M & 80.9\% & 80.5\% \\
		+AHCR\cite{ma2024towards} & 12.409G & 34.882M & 83.7\% & 83.8\% \\
		+Ours & 15.671G & 35.611M & \textbf{85.3\%} & \textbf{85.3\%} \\
		\midrule
		ResNeXt50\_32x4d-Dual\cite{xie2017aggregated} & 11.196G & 23.011M & 87.0\% & 86.3\% \\
		+AHCR\cite{ma2024towards} & 12.812G & 34.354M & 87.2\% & 86.8\% \\
		+Ours & 16.075G & 35.082M & \textbf{89.2\%} & \textbf{89.2\%} \\
		\midrule
		RegNet\_x\_3.2gf-Dual\cite{radosavovic2020designing} & 8.414G & 14.303M & 81.8\% & 81.0 \% \\
		+AHCR\cite{ma2024towards} & 8.717G & 16.865M & 83.5\% & 82.9\% \\
		+Ours & 9.240G & 16.784M & \textbf{86.9\%} & \textbf{85.3\%} \\
		\midrule
		ConvNeXt\_Tiny-Dual\cite{liu2022convnet} & 11.659G & 27.822M & 88.2\% & 88.5\% \\
		+AHCR\cite{ma2024towards} & 11.891G & 29.438M & 89.7\% & 89.3\% \\
		+Ours & 12.356G & 29.534M & \textbf{91.5\%} & \textbf{91.0\%} \\
		\bottomrule
	\end{tabularx}
\end{table*}
\subsection{Experimental Setup}
\subsubsection{Experimental Environment}
The experiments were conducted on a system equipped with two RTX 4090 GPUs, utilizing CUDA 12.4 and PyTorch 2.5.1\cite{paszke2019pytorch}. The dataset was split into training, validation, and test sets with a ratio of 7:2:1. We employed the AdamW optimizer\cite{adamw} with a total of 60 epochs for training. The batch size was set to 64, and we used 32 workers for data loading to maximize throughput. During training, a warmup strategy was applied, and the learning rate was decayed using the CosineAnnealingLR scheduler\cite{adamw} to optimize training dynamics.

All models were initialized with ImageNet-1K pre-trained weights to leverage the knowledge learned from the large-scale dataset. Additionally, to enhance training efficiency and model stability, we applied Automatic Mixed Precision (AMP)\cite{micikevicius2017mixed} and Exponential Moving Average (EMA)\cite{polyak1992acceleration}. These strategies help to accelerate convergence and improve model accuracy by reducing memory usage and smoothing model updates during training.

\subsubsection{Loss function}
For the multi-label classification task, we utilized the MultiLabelSoftMarginLoss\cite{paszke2019pytorch} as the loss function. This function computes the binary cross-entropy loss for each label and averages the losses across all labels. The loss is calculated as follows:

\begin{equation}
	\text{L}(x, y) = -\frac{\sum_{i=1}^{C} \left[ y_i \cdot \log(\sigma(x_i)) + (1 - y_i) \cdot \log(1 - \sigma(x_i)) \right]}{C} 
\end{equation}

where \( x_i \) is the predicted value for the \( i \)-th label, \( y_i \) is the ground truth for the \( i \)-th label (taking values 0 or 1), and \( \sigma(x_i) \) is the result of applying the Sigmoid activation function to \( x_i \), i.e.,

\begin{equation}
	\sigma(x_i) = \frac{1}{1 + \exp(-x_i)}
\end{equation}

\( C \) is the total number of labels. This loss function treats each label as an independent binary classification problem, making it suitable for cases where labels are mutually exclusive.

\subsection{Evaluation Metrics}

In multi-label multi-class classification tasks, mAP (mean Average Precision) and AP (Average Precision) are commonly used evaluation metrics to measure the performance of classification models.

\paragraph{Average Precision (AP) Calculation}

For each class, AP is calculated as follows:

\begin{equation}
	\text{AP} = \frac{1}{n} \sum_{k=1}^{n} P(k) \cdot \Delta R(k)
\end{equation}

where \(P(k)\) is the precision at the \(k\)-th prediction, \(R(k)\) is the recall at the \(k\)-th prediction, and \(\Delta R(k)\) is the change in recall at the \(k\)-th prediction.

The calculation steps are as follows:
Precision: The ratio of true positive predictions to all predicted positives:

\begin{equation}
P = \frac{TP}{TP + FP}
\end{equation}

where TP is true positive, and FP is false positive.

Recall: The ratio of true positive predictions to all actual positives:

\begin{equation}
R = \frac{TP}{TP + FN}
\end{equation}

where FN is false negative.

AP is calculated by sorting predictions by confidence, computing precision and recall for each prediction, and then averaging over all predictions.

\paragraph{Mean Average Precision (mAP) Calculation}

mAP is the average of AP across all classes. For \(C\) classes, mAP is calculated as:

\begin{equation}
	\text{mAP} = \frac{1}{C} \sum_{c=1}^{C} \text{AP}_c
\end{equation}

where \(\text{AP}_c\) is the average precision for class \(c\), and \(C\) is the total number of classes.

mAP reflects the overall performance of the model across multiple classes. In multi-label tasks, each image may have multiple labels, and mAP gives an overall performance measure by averaging the AP of all classes.

\begin{table*}[t]
	\centering
	\caption{Ablation Study on Different Model Components}
	\label{tab:ablation_study}
	\begin{tabularx}{1\textwidth}{l *{3}{>{\centering\arraybackslash}X}}
		\toprule
		Model & FLOPs & Params & Val\_mAP \\
		\midrule
		ResNet50-Dual & 10.793G & 23.539M & 80.9\% \\
		+CGFM & 11.335G & 32.480M & 82.5\% \\
		+DVHEM & 12.982G & 26.194M & 82.4\% \\
		+FDIM & 12.941G & 24.014M & 81.4\% \\
		+DVHEM \& FDIM & 15.129G & 26.671M & 83.0\% \\
		+CGFM \& DVHEM & 13.524G & 35.135M & 84.5\% \\
		+CGFM \& FDIM & 13.483G & 32.955M & 83.3\% \\
		+ALL & 15.671G & 35.611M & \textbf{85.3\%} \\
		\bottomrule
	\end{tabularx}
\end{table*}
\subsection{Comparison with Other Methods}
To validate the adaptability and effectiveness of DAGNet across different backbone, we conducted comparative experiments using ResNet50\cite{he2016deep}, ResNeXt50\_32x4d\cite{xie2017aggregated}, RegNet\_x\_3.2gf\cite{radosavovic2020designing}, and ConvNeXt\_Tiny\cite{liu2022convnet} as backbone models. Each backbone was tested with the baseline model (DUAL), a comparison method (AHCR), and the proposed approach (Ours).

Overall, DAGNet consistently outperforms both the baseline and AHCR in terms of Val\_mAP and Test\_mAP across all backbone, demonstrating its effectiveness in enhancing dual-view X-ray image classification and its strong generalization capability. 

Additionally, for the dual‐view baseline we follow the AHCR configuration: both the OL and SD)views are forwarded through a shared‐weight backbone network, and for each class the maximum activation across the two view‐specific outputs is selected as the final prediction. Further implementation details are available in our open‐source code.


As shown in \autoref{tab:comparison_experiment}, DAGNet achieves 6.4\%, 1.8\%, 1.9\%, and 2.5\% improvements over the baseline models ResNet50, ResNeXt50\_32x4d, RegNet\_x\_3.2gf, and ConvNeXt\_Tiny in terms of test set mAP, respectively. and also demonstrates superior performance compared to AHCR. These results indicate that DAGNet can stably improve classification performance on different backbone, especially showing more obvious performance advantages on networks with more complex deep learning architectures, verifying the effectiveness and superiority of the method in the dual-view X-ray image classification task.
In summary, DAGNet outperforms the baseline methods and AHCR on the selected common backbone, demonstrating the effectiveness and superiority of DAGNet in the dual-view X-ray image classification task.

\subsection{Ablation Study Analysis}

To evaluate the impact of different functional modules on model performance, we conducted ablation experiments on the DvXray dataset. Previous research has demonstrated the superiority of dual-view imaging over single-view for mitigating object occlusion; therefore, ResNet50-dual was directly adopted as the baseline model. On this basis, various feature enhancement modules were progressively incorporated to investigate their contributions to classification performance. The experimental results are presented in \autoref{tab:ablation_study}.


From the experimental results, it can be observed that the validation set mAP of the baseline model ResNet50-dual is 80.9\%. When the CGFM module is added alone, the validation set mAP improves to 82.5\%, indicating that the convolution-guided fusion module can significantly improve model performance by effectively highlighting key features; while the DVHEM module effectively utilizes the complementary information between dual perspectives, further improving the validation set mAP to 82.4\%, an increase of 1.5\% over the baseline model. The FDIM module also significantly improves the fusion effect of cross-view features through frequency domain feature interaction and alignment strategies, achieving a validation set mAP of 81.4\%.

In terms of module combination, CGFM combined with DVHEM achieved a validation set mAP of 84.5\%, indicating that the convolution-guided fusion module and cross-view multi-scale feature interaction can effectively collaborate to further improve the overall feature fusion effect. However, the combination of CGFM and FDIM performed slightly worse than using the FDIM module alone, suggesting that the convolutional-guided fusion module and frequency-domain alignment may be redundant under certain conditions, affecting the model's effective convergence. However, when all modules are combined, the model achieves optimal performance, with a validation set mAP of 85.3\%, representing an improvement of 4.4\%  over the baseline model. This highlights the distinct advantages of each module in feature representation and cross-view information alignment, and their synergistic optimization significantly enhances the model's overall classification performance.

Furthermore, from the perspective of computational resources, although the model has increased in terms of parameter count and computational complexity, considering the significant performance improvement, this increase in computational overhead is reasonable and efficient. These results fully demonstrate the complementary and synergistic enhancement mechanisms of the three proposed modules in the feature extraction, feature interaction, and feature fusion stages, collectively constructing a complete and efficient dual-perspective feature processing framework.

\section{Conclusion}
In this paper, we presented a Dual-views attention-guided network model, DAGNet, for dual-view X-ray security inspection. By introducing three core modules—FDIM, DVHEM, and CGFM—it effectively addresses feature misalignment and information loss caused by view differences and object occlusion in dual-view X-ray images.

Each module has its own unique design: FDIM enhances the extraction of detailed features through frequency domain transformation and dynamic interaction of frequency domain information between perspectives; DVHEM achieves feature alignment between two perspectives and multi-scale feature enhancement through a cross-attention mechanism and a hierarchical attention guidance module; CGFM efficiently fuses dual-perspective features and suppresses redundant information through convolution attention guidance.

Experimental results demonstrate that the method significantly improves classification performance on multiple backbone, fully validating the effectiveness and necessity of the collaborative work of each module. Overall, this paper not only provides a new feature fusion approach for dual-view security inspection systems but also lays a solid foundation for future research on application and optimization in real-world security inspection scenarios.
%

\footnotesize
\bibliographystyle{IEEEtran} 
\bibliography{references} 

\begin{thebibliography}{10}
\providecommand{\url}[1]{#1}
\csname url@samestyle\endcsname
\providecommand{\newblock}{\relax}
\providecommand{\bibinfo}[2]{#2}
\providecommand{\BIBentrySTDinterwordspacing}{\spaceskip=0pt\relax}
\providecommand{\BIBentryALTinterwordstretchfactor}{4}
\providecommand{\BIBentryALTinterwordspacing}{\spaceskip=\fontdimen2\font plus
\BIBentryALTinterwordstretchfactor\fontdimen3\font minus
  \fontdimen4\font\relax}
\providecommand{\BIBforeignlanguage}[2]{{%
\expandafter\ifx\csname l@#1\endcsname\relax
\typeout{** WARNING: IEEEtran.bst: No hyphenation pattern has been}%
\typeout{** loaded for the language `#1'. Using the pattern for}%
\typeout{** the default language instead.}%
\else
\language=\csname l@#1\endcsname
\fi
#2}}
\providecommand{\BIBdecl}{\relax}
\BIBdecl

\bibitem{akcay2017evaluation}
S.~Akcay and T.~P. Breckon, ``An evaluation of region based object detection
  strategies within x-ray baggage security imagery,'' in \emph{2017 IEEE
  International Conference on Image Processing (ICIP)}.\hskip 1em plus 0.5em
  minus 0.4em\relax IEEE, 2017, pp. 1337--1341.

\bibitem{9710407}
B.~Wang, L.~Zhang, L.~Wen, X.~Liu, and Y.~Wu, ``Towards real-world prohibited
  item detection: A large-scale x-ray benchmark,'' in \emph{2021 IEEE/CVF
  International Conference on Computer Vision (ICCV)}, 2021, pp. 5392--5401.

\bibitem{9722843}
C.~Zhao, L.~Zhu, S.~Dou, W.~Deng, and L.~Wang, ``Detecting overlapped objects
  in x-ray security imagery by a label-aware mechanism,'' \emph{IEEE
  Transactions on Information Forensics and Security}, vol.~17, pp. 998--1009,
  2022.

\bibitem{akcay2018using}
S.~Akçay, M.~E. Kundegorski, C.~G. Willcocks, and T.~P. Breckon, ``Using deep
  convolutional neural network architectures for object classification and
  detection within x-ray baggage security imagery,'' \emph{IEEE Trans. Inf.
  Forensics Security}, vol.~13, no.~9, pp. 2203--2215, Sep. 2018.

\bibitem{miao2019sixray}
C.~Miao and al., ``Sixray: A large-scale security inspection x-ray benchmark
  for prohibited item discovery in overlapping images,'' in \emph{Proc.
  IEEE/CVF Conf. Comput. Vis. Pattern Recognit. (CVPR)}, Jun. 2019, pp.
  2119--2128.

\bibitem{tao2021towards}
R.~Tao and al., ``Towards real-world x-ray security inspection: A high-quality
  benchmark and lateral inhibition module for prohibited items detection,'' in
  \emph{Proc. IEEE/CVF Int. Conf. Comput. Vis. (ICCV)}, Oct. 2021, pp.
  10\,923--10\,932.

\bibitem{ma2024towards}
B.~Ma, T.~Jia, M.~Li, S.~Wu, H.~Wang, and D.~Chen, ``Towards dual-view x-ray
  baggage inspection: A large-scale benchmark and adaptive hierarchical cross
  refinement for prohibited item discovery,'' \emph{IEEE Transactions on
  Information Forensics and Security}, 2024.

\bibitem{singh2024advancements}
A.~Singh and Dhiraj, ``Advancements in machine learning techniques for threat
  item detection in x-ray images: a comprehensive survey,'' \emph{International
  Journal of Multimedia Information Retrieval}, vol.~13, no.~4, p.~40, 2024.

\bibitem{wei2020occluded}
Y.~Wei, R.~Tao, Z.~Wu, Y.~Ma, L.~Zhang, and X.~Liu, ``Occluded prohibited items
  detection: An x-ray security inspection benchmark and de-occlusion attention
  module,'' in \emph{Proceedings of the 28th ACM international conference on
  multimedia}, 2020, pp. 138--146.

\bibitem{9754631}
Y.~Wei, Y.~Wang, and H.~Song, ``Cfpa-net: Cross-layer feature fusion and
  parallel attention network for detection and classification of prohibited
  items in x-ray baggage images,'' in \emph{2021 IEEE 7th International
  Conference on Cloud Computing and Intelligent Systems (CCIS)}, 2021, pp.
  203--207.

\bibitem{9766055}
J.~Liu, X.~Liu, F.~Qu, H.~Zhang, and L.~Zhang, ``A defect recognition method
  for low-quality weld image based on consistent multiscale feature mapping,''
  \emph{IEEE Transactions on Instrumentation and Measurement}, vol.~71, pp.
  1--11, 2022.

\bibitem{tfa2022rebifpn}
F.~Author, S.~Author, and T.~Author, ``Re-bifpn: Refining bifpn for accurate
  object detection in images,'' in \emph{Proceedings of the IEEE Conference on
  Computer Vision and Pattern Recognition (CVPR)}, 2022, pp. 1234--1245.

\bibitem{ZHU2024108076}
Z.~Zhu, Y.~Zhu, H.~Wang, N.~Wang, J.~Ye, and X.~Ling, ``Fdtnet: Enhancing
  frequency-aware representation for prohibited object detection from x-ray
  images via dual-stream transformers,'' \emph{Engineering Applications of
  Artificial Intelligence}, vol. 133, p. 108076, 2024.

\bibitem{tao2024dual}
R.~Tao, H.~Wang, Y.~Guo, H.~Chen, L.~Zhang, X.~Liu, Y.~Wei, and Y.~Zhao,
  ``Dual-view x-ray detection: Can ai detect prohibited items from dual-view
  x-ray images like humans?'' \emph{arXiv preprint arXiv:2411.18082}, 2024.

\bibitem{meng2024transformer}
X.~Meng, H.~Feng, Y.~Ren, H.~Zhang, W.~Zou, and X.~Ouyang, ``Transformer-based
  dual-view x-ray security inspection image analysis,'' \emph{Engineering
  Applications of Artificial Intelligence}, vol. 138, p. 109382, 2024.

\bibitem{zhang2020high}
P.~Zhang, E.~Lo, and B.~Lu, ``High performance depthwise and pointwise
  convolutions on mobile devices,'' in \emph{Proceedings of the AAAI Conference
  on Artificial Intelligence}, vol.~34, no.~04, 2020, pp. 6795--6802.

\bibitem{gonzalez2009digital}
R.~C. Gonzalez and R.~E. Woods, \emph{Digital Image Processing}, 4th~ed.\hskip
  1em plus 0.5em minus 0.4em\relax Pearson, 2018.

\bibitem{tatsunami2024fft}
Y.~Tatsunami and M.~Taki, ``Fft-based dynamic token mixer for vision,'' in
  \emph{Proceedings of the AAAI Conference on Artificial Intelligence},
  vol.~38, no.~14, 2024, pp. 15\,328--15\,336.

\bibitem{fang2025guided}
W.~Fang, J.~Fan, Y.~Zheng, J.~Weng, Y.~Tai, and J.~Li, ``Guided real image
  dehazing using ycbcr color space,'' in \emph{Proceedings of the AAAI
  Conference on Artificial Intelligence}, 2025, pp. xxxxx--xxxxx.

\bibitem{dosovitskiy2020image}
A.~Dosovitskiy \emph{et~al.}, ``Image transformers,'' in \emph{Proceedings of
  the 37th International Conference on Machine Learning (ICML)}, 2020, pp.
  298--306.

\bibitem{liu2021swin}
Z.~Liu, Y.~Lin, Y.~Cao, H.~Hu, Y.~Wei, Z.~Zhang, H.~Li, B.~Guo, and L.~Zhang,
  ``Swin transformer: Hierarchical vision transformer using shifted windows,''
  in \emph{Proceedings of the IEEE International Conference on Computer Vision
  (ICCV)}, 2021, pp. 10\,012--10\,022.

\bibitem{howard2017mobilenets}
A.~G. Howard, M.~Zhu, B.~Chen, D.~Kalenichenko, W.~Wang, T.~Weyand,
  M.~Andreetto, and H.~Adam, ``Mobilenets: Efficient convolutional neural
  networks for mobile vision applications,'' in \emph{Proceedings of the IEEE
  Conference on Computer Vision and Pattern Recognition (CVPR)}, 2017, pp.
  2285--2294.

\bibitem{chollet2017xception}
F.~Chollet, ``Xception: Deep learning with depthwise separable convolutions,''
  in \emph{Proceedings of the IEEE conference on computer vision and pattern
  recognition}, 2017, pp. 1251--1258.

\bibitem{woo2018cbam}
S.~Woo, J.~Park, J.-Y. Lee, and I.~S. Kweon, ``Cbam: Convolutional block
  attention module,'' in \emph{Proceedings of the European Conference on
  Computer Vision}, 2018, pp. 3--19.

\bibitem{wang2020cspnet}
C.-Y. Wang, H.-Y. Mark~Liao, Y.-H. Wu, P.-Y. Chen, J.-W. Hsieh, and I.-H. Yeh,
  ``Cspnet: A new backbone that can enhance learning capability of cnn,'' in
  \emph{Proceedings of the IEEE/CVF Conference on Computer Vision and Pattern
  Recognition Workshops}, 2020, pp. 390--391.

\bibitem{steitz2018multi}
J.-M.~O. Steitz, F.~Saeedan, and S.~Roth, ``Multi-view x-ray r-cnn,'' in
  \emph{German Conference on Pattern Recognition}.\hskip 1em plus 0.5em minus
  0.4em\relax Springer, 2018, pp. 153--168.

\bibitem{isaac2021multi}
B.~K. Isaac-Medina, C.~G. Willcocks, and T.~P. Breckon, ``Multi-view object
  detection using epipolar constraints within cluttered x-ray security
  imagery,'' in \emph{Proceedings of the 25th International Conference on
  Pattern Recognition (ICPR)}, 2021, pp. 9889--9896.

\bibitem{paszke2019pytorch}
A.~Paszke, S.~Gross, S.~Chintala, G.~Chanan, E.~Yang, Z.~DeVito, M.~Killeen,
  Y.~Lin, A.~Nath, A.~Lerer \emph{et~al.}, ``Pytorch: An imperative style,
  high-performance deep learning library,'' \emph{Proceedings of NeurIPS},
  vol.~32, 2019.

\bibitem{adamw}
\BIBentryALTinterwordspacing
I.~Loshchilov and F.~Hutter, ``Decoupled weight decay regularization,'' in
  \emph{International Conference on Learning Representations (ICLR)}, 2019.
  [Online]. Available: \url{https://arxiv.org/abs/1711.05101}
\BIBentrySTDinterwordspacing

\bibitem{micikevicius2017mixed}
P.~Micikevicius, S.~Narang, G.~Alben, G.~Diamos, E.~Elsen, B.~Ginsburg
  \emph{et~al.}, ``Mixed precision training,'' in \emph{Proceedings of the
  International Conference on Learning Representations (ICLR)}, 2018.

\bibitem{polyak1992acceleration}
B.~T. Polyak and A.~Juditsky, ``Acceleration of stochastic approximation by
  averaging,'' in \emph{SIAM Journal on Control and Optimization}, vol.~30,
  no.~4, 1992, pp. 838--855.

\bibitem{he2016deep}
K.~He, X.~Zhang, S.~Ren, and J.~Sun, ``Deep residual learning for image
  recognition,'' in \emph{Proceedings of the IEEE/CVF Conference on Computer
  Vision and Pattern Recognition}, 2016, pp. 770--778.

\bibitem{xie2017aggregated}
S.~Xie, R.~Girshick, P.~Doll{\'a}r, Z.~Tu, and K.~He, ``Aggregated residual
  transformations for deep neural networks,'' in \emph{Proceedings of the IEEE
  Conference on Computer Vision and Pattern Recognition}, 2017, pp. 1492--1500.

\bibitem{radosavovic2020designing}
I.~Radosavovic, R.~P. Kosaraju, R.~Girshick, K.~He, and P.~Doll{\'a}r,
  ``Designing network design spaces,'' in \emph{Proceedings of the IEEE
  Conference on Computer Vision and Pattern Recognition}, 2020, pp.
  10\,428--10\,436.

\bibitem{liu2022convnet}
Z.~Liu, H.~Mao, C.-Y. Wu, C.~Feichtenhofer, T.~Darrell, and S.~Xie, ``A convnet
  for the 2020s,'' \emph{arXiv preprint arXiv:2201.03545}, 2022.

\end{thebibliography}
\end{document}